\documentclass[a4paper, 10pt, conference]{ieeeconf}      
\usepackage{FG2021}
\usepackage[utf8]{inputenc}
\usepackage{graphicx}
\usepackage{amsmath}
\usepackage{caption}
\usepackage{bbding}
\usepackage{amssymb}
\usepackage{subcaption}
\FGfinalcopy 

\IEEEoverridecommandlockouts                              
\def\FGPaperID{174} 

\title{\LARGE \bf
MD-CSDNetwork: Multi-Domain Cross Stitched Network for Deepfake Detection
}

\author{\parbox{16cm}{\centering
    {\large Aayushi Agarwal$^1$, Akshay Agarwal$^2$, Sayan Sinha$^3$, Mayank Vatsa$^1$, and Richa Singh$^1$}\\
    {\normalsize
    $^1$IIT Jodhpur, India,
    $^2$University at Buffalo, USA, and 
    $^3$IIT Kharagpur} \\
    {\normalsize
    aayushi.agarwal007@gmail.com,
    aa298@buffalo.edu, sayan.sinha@iitkgp.ac.in, and \{richa, mvatsa\}@iitj.ac.in}}
}
\usepackage{booktabs, multirow, graphicx, tabularx, subcaption}

\begin{document}

\ifFGfinal
\thispagestyle{empty}
\pagestyle{empty}
\else
\author{Anonymous FG2021 submission \\}
\pagestyle{plain}
\fi
\maketitle

\begin{figure*}[htp]
    \centering
    \includegraphics[width=14cm]{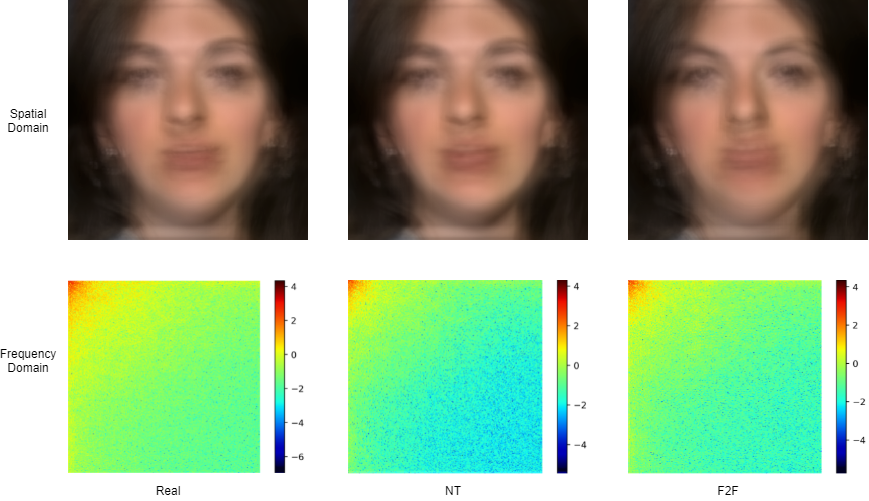}
    \caption{Comparison of DCT-based frequency spectrum between a real video and a manipulated video. From left to right, row 1 shows the images obtained for a video by averaging the pixel values of all the frames for a Real, Neural Textures \cite{10.1145/3306346.3323035} and Face2Face \cite{10.1145/3292039} videos. There are no observable differences seen in all the three images. From left to right, row 2 shows the heatmaps for the frequency spectrum obtained after averaging the log-scaled DCT coefficients of all the frames for a Real, Neural Textures (NT), and Face2Face (F2F) video. We clearly see the differences in color variations as we move from the low-frequency (top-left) to the high-frequency (bottom-right) region between real vs manipulated videos. Best viewed in color.}
    \label{fig:fig1}
\end{figure*}

\begin{abstract}
The rapid progress in the ease of creating and spreading ultra-realistic media over social platforms calls for an urgent need to develop a generalizable deepfake detection technique. It has been observed that current deepfake generation methods leave discriminative artifacts in the frequency spectrum of fake images and videos. Inspired by this observation, in this paper, we present a novel approach, termed as MD-CSDNetwork, for combining the features in the spatial and frequency domains to mine a shared discriminative representation for classifying \textit{deepfakes}. MD-CSDNetwork is a novel cross-stitched network with two parallel branches carrying the spatial and frequency information, respectively. We hypothesize that these multi-domain input data streams can be considered as related supervisory signals. The supervision from both branches ensures better performance and generalization. Further, the concept of cross-stitch connections is utilized where they are inserted between the two branches to learn an optimal combination of domain-specific and shared representations from other domains automatically. Extensive experiments are conducted on the popular benchmark dataset namely FaceForeniscs++ for forgery classification. We report improvements over all the manipulation types in FaceForensics++ dataset and comparable results with state-of-the-art methods for cross-database evaluation on the Celeb-DF dataset and the Deepfake Detection Dataset.

\end{abstract}

\section{INTRODUCTION}

Manipulating videos or images is not new; earlier seamless manipulation used to be a skillful task requiring a lot of resources, time, and dexterous artists. However, recent growth in the deep learning technology has given anyone the ability to make a convincing fake video, which can also be misused by some people to “weaponize” it for political or other malicious purposes. A Reddit user by the name of ‘deepfakes’ using deep learning to make pornographic videos and posting them online, President Obama using an expletive to describe President Trump and Mark Zuckerberg admitting that Facebook's true goal is to manipulate and exploit its users are some notable examples of such misuse \cite{deepfakes-2, deepfakes-1}. Audios can be \textit{deepfaked} as well to create “voice skins'' or “voice clones'' of public figures. A recent example where audio deepfakes have been exploited was a phone call by a fraudster mimicking the voice of a German energy firm's CEO. The chief of a UK subsidiary of the firm was tricked, and they paid nearly £200,000 into a Hungarian bank account \cite{deepfakes-1}. Similar scams have reportedly used recorded WhatsApp voice messages. All these instances suggest that developing robust and efficient methods of deepfake detection is of paramount importance.

Most of the existing \textit{deepfakes} alter facial content, and these manipulations can be categorized in four broad groups depending upon the levels of face manipulation \cite{tolosana2020deepfakes}: 

\begin{itemize}
    \item identity swap where a face from a source video replaces a face in the target video usually employing computer-graphics or deep learning based methods, 
    \item facial expression reenactment where the expressions in the source video are transferred to the target video while maintaining the identity of the target person,
    \item complete synthesis of high-quality and realistic new faces using generative adversarial networks (GANs). This is different from manipulating only a certain part (mostly facial region) in the image, and  
    \item attribute manipulation where various attributes such as skin color and hair color are edited. 
\end{itemize}

\subsection{Literature Review}
A wide variety of approaches have been proposed for verifying the authenticity of images and videos in the vast area of digital media forensics. Previous research directions have used physiological signals like eye blinking \cite{li2018ictu}, inconsistent head poses \cite{8683164}, and biological signals not preserved in fake videos \cite{Ciftci_2020} as well as phoneme-viseme mismatches in videos \cite{9151013} as the basis for detecting deepfake content. Various deep learning based detection methods \cite{8630761,Agarwal_2019_CVPR_Workshops,cozzolino2019forensictransfer,9150778,li2019exposing,wang2020fakespotter,8953774,8014963} have been proposed to mitigate the risks of deepfakes. Some notable examples include \cite{9360904}, where the combination of a static biometric for face recognition along with a temporal \& behavioral biometric is used for face-swap deepfakes. Also, Face X-Ray \cite{Li_2020_CVPR} uses a more generic approach for detection by identifying whether the input image can be decomposed into a blending of different images or not. \cite{8682602} uses a capsule network to detect various kinds of deepfake attacks created using deep learning. Further, \cite{Davis_2017_CVPR_Workshops} uses a face classification stream and a patch triplet stream for leveraging features capturing local noise residuals and camera characteristics. A two-stream network is proposed for detecting image manipulations where one of the streams uses RGB image for identifying high-level tampering artifacts like contrast difference and unnatural boundaries \cite{8578214}. More recently, SSTNet \cite{9053969} uses a detection framework to detect tampered faces through spatial, steganalysis, and temporal features. Some methods have also used the artifacts produced in the frequency spectrum to mine better discriminative feature representations. $F^{3}$-Net \cite{qian2020thinking} have proposed a two-stream collaborative learning framework that makes use of frequency-aware decomposed image components and local frequency statistics to mine better forgery patterns. Masi \emph{et al.} \cite{masi2020twobranch} have presented a two-branched network structure with one branch propagating the original information in the color domain while the other suppressing the face content and amplifying multi-band frequencies using a Laplacian of Gaussian (LoG). Some recent works \cite{8695364,Wang_2020_CVPR,9010964} focus on identifying special artificial fingerprints common across all GAN synthesized images to detect fake images from their real counterparts. \cite{8639163,masi2020twobranch,Sabir_2019_CVPR_Workshops,trinh2021interpretable} have also exploited the temporal information between the frames in a video to isolate deepfakes through sequence-based models.

Most of the current deepfake creation methods, especially the ones which use generative models (e.g. GANs) to create manipulated faces or completely new synthetic images, rely on convolution based upsampling methods due to which they fail to reproduce the spectral distributions of natural images \cite{Durall_2020_CVPR}. This creates artifacts in the frequency spectrum of fake images, which can be used as discriminatory evidence against the real ones. The frequency artifacts are more common and generic across a variety of manipulations, including compression scenarios, where the information in the spatial domain is severely affected. 

Recent approaches \cite{durall2020unmasking,frank2020leveraging,9035107} specifically focus on identifying images synthesized through GANs by using the frequency-related features as inputs to a classifier. In particular, \cite{durall2020unmasking, frank2020leveraging} have proposed spectrum-based classifiers that input the frequency features rather than pixel input for detecting synthetically generated images. However, such a classifier may not be suitable for detecting other kinds of facial manipulations, for example, the manipulations in the widely used FaceForensics++ (FF++) \cite{Rossler_2019_ICCV} dataset. Here, the artifacts in the spatial domain are also essential and cannot be neglected completely. On the other hand, Face X-Ray \cite{Li_2020_CVPR} proposed a novel image representation to identify any blending boundary artifacts introduced during the manipulation process but uses only spatial domain information. This may not be sufficient for detecting GAN-based manipulations or highly realistic synthetic images that might leave comparatively much fewer artifacts in the spatial domain. Therefore, a combination of features from both the domains is more useful to detect various kinds of manipulation types. 


\subsection{Research Contributions}
In this paper, we propose a novel approach, termed as MD-CSDNetwork, of combining information from the two domains, spatial and frequency, by learning a shared representation at both low-level and high-level feature maps. We draw an analogy that features from spatial and frequency domains can act as two related supervisory signals similar to related tasks in a multi-task learning problem. We jointly learn the multi-domain feature representation via the proposed cross-stitched network architecture for learning discriminative features. Specifically, each cross-stitch unit learns an optimal linear combination of domain-specific and shared representations from other domains at various levels in the network. Depending upon the input manipulation type, the network learns the weights of the activation maps from both the domains as required. In other words, the proposed approach enables learning the contribution of features from both the domains, and the training signals from the two domains help each other to enhance the performance and generalizability. The major contributions of this paper are:


 \textit{We propose a cross-stitched network architecture for jointly learning features from both spatial and frequency domain information. It gives the network the flexibility to combine features from both the domains at various levels depending upon the input data. We hypothesize that the process of learning a shared representation from two different domain features of the same image can be considered similar to learning a shared representation from multiple related tasks in a multi-task learning setting. Such a representation enables supervision from both feature domains, which helps the model to learn better discriminative representations. We have experimentally evaluated the use of Discrete Cosine Transform (DCT) spectrum over Discrete Wavelet Transform (DWT) or Fast Fourier Transform (FFT) for extracting the frequency-related features in the proposed network. Through extensive experimentation, we show that the proposed model with DCT achieves a considerable improvement in the detection performance.}



\begin{figure*}[htp]
    \centering
    \includegraphics[width=15cm]{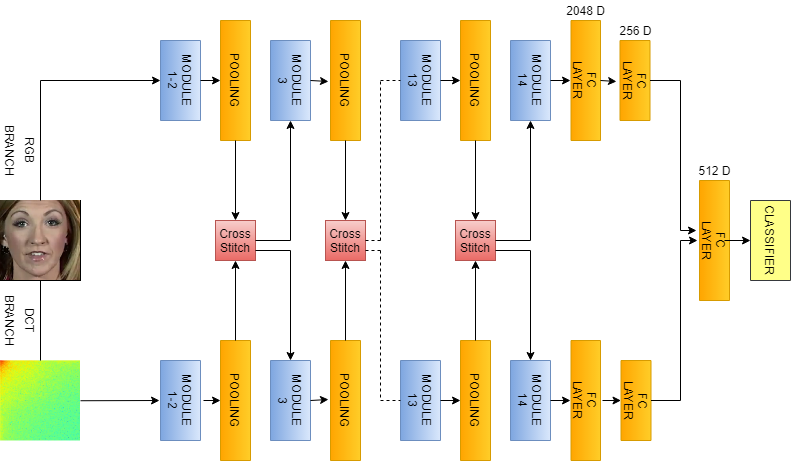}
    \caption{Network Architecture for the proposed method. The two parallel branches carry the spatial and frequency information, respectively. The top branch inputs the pixel values, and the bottom branch inputs the log-scaled DCT spectrum coefficients. The cross-stitch \cite{Misra_2016_CVPR} units are applied after the pooling layers in the backbone XceptionNet \cite{Chollet_2017_CVPR}. Multiple such units have been represented as dotted lines. In MD-CSDNetwork, four cross-stitch units have been used.}
    \label{fig:fig2}
\end{figure*}

\section{Proposed Approach}
\subsection{Motivation}
To understand deepfake artifacts in the frequency domain, Fig. \ref{fig:fig1} shows plots of the DCT coefficient spectrum of real and fake images for two manipulation types namely Neural Textures (NT) \& Face2Face (F2F) in FF++. The horizontal direction corresponds to frequencies in the $x$-direction, and the vertical direction corresponds to frequencies in the $y$-direction. The frequencies increase as we move from left to right and top to bottom for $x$ and $y$ directions, respectively. 
We observe clear distinguishability between the spectrums of real and fake images, especially in the high-frequency regions, which implies that deepfakes are unable to mimic the real spectrum variations.


We further draw the motivation for combining information from spatial and frequency domains by formulating the learning procedure similar to a multi-task learning problem. In multi-task learning, related tasks share the representations learned from multiple supervisory tasks with direct supervision from the primary task and indirect supervision from auxiliary related tasks. This improves generalization by leveraging the domain specific information contained in the training signals of the related tasks. Misra \emph{et al.} \cite{Misra_2016_CVPR} explains that deciding how much sharing one needs to enforce among multiple tasks is very cumbersome and highly dependent on the tasks at hand as well as the input data. There can be a spectrum of possible network architectures ranging from all the layers (except the last layers) being shared to no sharing at all. They provide a principled approach to mitigate this problem with the use of cross-stitch units which learn an optimal combination of both task-specific and shared representations automatically. Hence, the tasks supervise how much sharing is needed for improved classification.

Inspired by the work done in \cite{Misra_2016_CVPR}, we hypothesize that two input streams, one carrying the spatial information of an image and the other carrying the frequency information of the same image, can be considered as two related supervisory signals. We devise a cross-stitched network that jointly learns a feature representation from both the domains with parallel branches carrying the two input data streams and connected with multiple cross-stitch units at various levels. These units enable direct supervision from themselves and indirect supervision from the other branch and inadvertently allow information flow between both the domains. The cross-stitch units learn an optimal combination of domain-specific and shared representations from other domains for better classification depending upon the input data. 

\subsection{Proposed MD-CSDNetwork Architecture}
As shown in Fig. \ref{fig:fig2}, the architecture of the proposed method consists of two parallel branches with cross-stitch units inserted after the max-pooling layers. These cross-connections applied at various levels in the network allow sharing between both the domains for both low and high-level semantics. We utilise four cross-stitch units in the proposed MD-CSDNetwork. XceptionNet \cite{Chollet_2017_CVPR} is used as a backbone for both the branches. The last fully connected layers are concatenated and connected to a classifier. The first branch inputs pixel features in the RGB domain, and the second inputs the DCT spectrum coefficients in the frequency domain. Both the branches are separately normalized. We next present the two ingredients of the proposed network: cross-stitch unit and DCT spectrum. 

\subsubsection{Cross-Stitch Unit}
Given two activation maps $x_{R}$ and $x_{D}$ obtained at a certain layer from the RGB and DCT branches respectively, we learn the linear combinations $x^{'}_{R}$ and $x^{'}_{D}$ of both the input activation maps which are then fed as inputs to the next layers (Eq. 1). This linear combination is parameterized using $\alpha$. Specifically, at any location $(i, j)$ in the activation map:\\
\begin{equation}
\begin{bmatrix}
$$x^{'ij}_{R}$$\\
\\
$$x^{'ij}_{D}$$
\end{bmatrix}
=
\begin{bmatrix}
$$\alpha_{RR}$$ & $$\alpha_{RD}$$\\
$$\alpha_{DR}$$ & $$\alpha_{DD}$$
\end{bmatrix}
\begin{bmatrix}
$$x^{ij}_{R}$$\\
\\
$$x^{ij}_{D}$$
\end{bmatrix}
\end{equation}

\noindent $\alpha_{RR}$, $\alpha_{DD}$ provides the weights to the activations of the same branch ($RGB$ and $DCT$, respectively), and $\alpha_{RD}$, $\alpha_{DR}$ provides the weights to the activations of the other branch. The network can decide to make $\alpha_{RD}$, $\alpha_{DR}$ zero for no sharing at all at a certain layer or choose a more shared representation by assigning higher values to these parameters. Their partial derivatives for loss $L$ with tasks $RGB$ and $DCT$ are:

\begin{equation}
\begin{bmatrix}
$$\frac{\partial L}{\partial x^{ij}_{R}}$$ \\
\\
$$\frac{\partial L}{\partial x^{ij}_{D}}$$
\end{bmatrix}
=
\begin{bmatrix}
$$\alpha_{RR}$$ & $$\alpha_{DR}$$\\
$$\alpha_{RD}$$ & $$\alpha_{DD}$$
\end{bmatrix}
\begin{bmatrix}
$$\frac{\partial L}{\partial x^{'ij}_{R}}$$ \\
\\
$$\frac{\partial L}{\partial x^{'ij}_{D}}$$
\end{bmatrix}
\end{equation}

\begin{equation}
   \frac{\partial L}{\partial \alpha_{RD}} = 
   \frac{\partial L}{\partial x^{'ij}_{D}} x^{ij}_{R}; \frac{\partial L}{\partial \alpha_{RR}} = \frac{\partial L}{\partial x^{'ij}_{R}} x^{ij}_{R} 
\end{equation}

\subsubsection{DCT Spectrum}
The magnitude of the DCT coefficients signifies the contribution of a certain frequency in the image. We specifically compute a 2-D DCT as a product of 1-D DCTs across the rows and then across the columns separately for all the channels in the input image. DCT \cite{1672377} has been widely used in image processing applications, especially in lossy compression, due to its excellent energy compaction properties. Due to this, the coefficients for high frequencies are much lower than low frequencies for natural images, and therefore we apply a natural logarithm to bring the coefficients at the same scale. The formula for 2-dimensional DCT is:
\\
\begin{equation}
\begin{split}
DCT(i, j) = & \frac{1}{\sqrt{2N}} C(i) C(j)
 \sum_{a = 0}^{N - 1} \sum_{b = 0}^{N - 1} pixel(a, b) \\ &
 \times cos[\frac{(2a + 1) i\pi}{2N}] cos[\frac{(2b + 1) i\pi}{2N}] \\ & C(x) = \frac{1}{\sqrt{2}} \ if\ x = 0,\ else\ 1\ if\ x > 0
\end{split}
\end{equation}

\subsection{Implementation Details}

We initialize the spatial branch with ImageNet \cite{5206848} weights and the frequency branch with random weights. For all the cross stitches, the same branch parameters ($\alpha_{RR}$, $\alpha_{DD}$) are initialized with 0.9 and different branch parameters ($\alpha_{RD}$, $\alpha_{DR}$) with 0.1 to ensure a convex linear combination of the parameter values and to make output activation maps of the same order in magnitude as the input activation maps. We train with a base learning rate of 2e-4 and use a slightly higher learning rate of 1e-3 for the cross-stitches. The learning rate is reduced by a factor of 0.2 every time when the validation loss does not decrease for three consecutive epochs. The batch size used is 32, and the model with the highest validation accuracy is used for testing. We use Adam optimizer \cite{kingma2017adam} with default values for the moments ($\beta 1$ = 0.9, $\beta 2$ = 0.999, $\epsilon$ = $10^{-8}$) and the whole model is trained end-to-end with a cross-entropy loss for binary classification. The network is trained with three V100 GPUs in DGX2 Server under TensorFlow programming environment. At an average, using this computational environment, a single training epoch on FF++ database takes about 4000 seconds.

\begin{figure*}[htp]
    \centering
    \includegraphics[width=15cm]{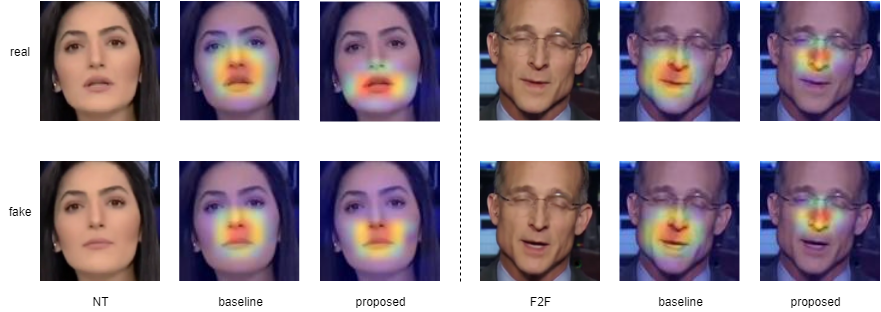}
    \caption{The Grad-CAM visualization of baseline XceptionNet \cite{Chollet_2017_CVPR} and the proposed method for two different manipulations in FF++ \cite{Rossler_2019_ICCV}. Best viewed in color.}
    \label{fig:fig3}
\end{figure*}

\section{EXPERIMENTS}

\subsection{Datasets and Protocol}

For performance evaluation, we use three benchmark datasets, namely FaceForensics++ (FF++) \cite{Rossler_2019_ICCV}, Celeb-DF \cite{9156368}, and Deepfake Detection (DFD) \cite{deepfakedetection}. 

\textbf{(a) FF++} consists of 1000 real videos collected from YouTube and 1000 fake videos for each of the four manipulation types, i.e. FaceSwap \cite{faceswap}, Face2Face \cite{10.1145/3292039}, Deepfakes \cite{deepfakesDataset} and Neural Textures \cite{10.1145/3306346.3323035}. To better simulate real-world scenarios, these videos are also compressed using the H.264 codec to light compression (c23) and high compression (c40) levels. All the videos contain full frontal faces without any occlusions. We used the freely available face detector dlib \cite{10.5555/1577069.1755843} to extract the largest face in a frame and used an enlarged face crop with a scale factor of 1.3. We used 720 videos for training and 140 videos for both validation and testing. We extracted 270 consecutive frames per video for training and 100 consecutive frames per video for validation and testing as per the settings in \cite{Rossler_2019_ICCV}. We perform two kinds of experiments: (i) train and test on all manipulation types taken together (ii) train and test on specific manipulation types for both low and high compression levels. For training with all the manipulation types, we balance the number of frames for real and fake videos accordingly. 

\textbf{(b) Celeb-DF} consists of 5639 fake videos and 590 real videos and \textbf{(c) DFD} which consists of 3068 fake videos and 363 real videos are recent high-quality datasets used for evaluating cross-database generalisation.

\subsection{Evaluation Metrics}
For all the experiments, we used the Accuracy (Acc) and the Area Under Receiver Operating Characteristic Curve (AUC) as the evaluation metrics. \textbf{(a) Acc.} Following FF++ \cite{Rossler_2019_ICCV}, we use the accuracy metric as the major evaluation metric in all our experiments. It is the most widely used metric in the face forgery detection task. \textbf{(b) AUC.} Following the recent works \cite{Li_2020_CVPR,9156368,masi2020twobranch,qian2020thinking}, we use this metric to evaluate the cross-data performance and ablation experiments. For better comparison with recent methods \cite{liu2021spatialphase,masi2020twobranch,qian2020thinking,9053969}, we report the average metric scores for all the frames in a video. For cross-database evaluation on Celeb-DF \cite{9156368}, we extract 25 frames per video and report the frame-level AUC scores as suggested in the original paper.

\begin{table*}[!t]
\centering
\scriptsize
\caption{\label{tab:tab1} Quantitative results (Acc \%) for both high-quality HQ (or low compression c23) \& low-quality LQ (or high compression c40) settings on all four manipulation types i.e FaceSwap (FS) \cite{faceswap}, Face2Face (F2F) \cite{10.1145/3292039}, Deepfakes (DF) \cite{deepfakesDataset} and Neural Textures (NT) \cite{10.1145/3306346.3323035} in the FF++ \cite{Rossler_2019_ICCV} dataset.}
\begin{tabular}{l|cc|cc|cc|ccc}\toprule
\multirow{2}{*}{Methods} &\multicolumn{2}{c|}{FS} &\multicolumn{2}{c|}{F2F} &\multicolumn{2}{c|}{DF} &\multicolumn{2}{c}{NT} \\\cmidrule{2-9}
&HQ &LQ &HQ &LQ &HQ &LQ &HQ &LQ \\\midrule
Steg. Features \cite{6197267} &- &68.93 &- &73.72 &- &73.64 &- &63.33 \\
Cozzolino \emph{et al.} \cite{10.1145/3082031.3083247} &- &73.79 &- &67.88 &- &85.45 &- &78.00 \\
Rahmouni \emph{et al.} \cite{8267647} &- &56.31 &- &64.23 &- &85.45 &- &60.07 \\
Bayar and Stamm \cite{10.1145/2909827.2930786} &- &82.52 &- &73.72 &- &84.55 &- &70.67 \\
MesoNET \cite{8630761} &- &61.17 &- &56.20 &- &87.27 &- &40.67 \\
SRMNet \cite{8578214} &- &89.10 &- &92.70 &- &91.90 &- &- \\
FF++ \cite{Rossler_2019_ICCV} &\underline{98.23} &92.17 &\underline{98.36} &88.95 &\underline{97.51}. &93.46 &\underline{93.59} &78.89 \\
SSTNet \cite{9053969} &- &94.09 &- &90.48 &- &95.33 &- &- \\
SPSL \cite{liu2021spatialphase} &- &92.26 &- &86.02 &- &93.48 &- &76.78 \\
$F^{3}$-Net \cite{qian2020thinking} &- &\textbf{96.53} &- &\textbf{95.32} &- &\textbf{97.97} &- &\textbf{83.32} \\
\noalign{\vskip 0.2mm}
\hline
\noalign{\vskip 1mm}
Proposed MD-CSDNetwork &\textbf{99.10} &\underline{94.64} &\textbf{99.19} &\underline{93.57} &\textbf{98.82} &\underline{97.34} &\textbf{94.55} &\underline{81.78} \\
\bottomrule
\end{tabular}

\end{table*}

\subsection{Results}
In this section, we discuss the evaluation results for the proposed model and compare it with the recent methods in deepfake detection\footnote{The two best results have been marked in the tables with bold face and underline.}.

\subsubsection{Evaluation on FF++}
Tables 1 and 2 show the comparison of the results of the proposed method on the test dataset with other recent methods on FF++ \cite{Rossler_2019_ICCV} dataset. For both the experiments, significant improvements in both AUC and accuracy metrics on high-quality HQ (or low-compression c23) levels is observed over the baseline XceptionNet. This shows that a combination of both spatial and frequency domain features can yield better results even for compressed videos, thus implying that frequency artifacts are robust to compression. For experiments with all manipulation types together, the proposed method achieves either comparable or better results when compared with the recent state-of-the-art methods such as \cite{liu2021spatialphase,masi2020twobranch,qian2020thinking,DBLP:conf/icml/TanL19} on low compression. We observe comparatively lesser results for the high compression scenario, which can be attributed to not exploiting the inter-frame temporal artifacts within a video or patch-based artifacts. 

For experiments on a single manipulation type, the network performs better than recent methods such as SSTNet \cite{9053969} which extracts spatial, steganalysis, and temporal features for detecting tampered videos, SRMNet \cite{8578214} which uses noise features extracted from SRM filters and comparable results with state-of-the-art results by $F^{3}$-Net \cite{qian2020thinking}. Among all four manipulation types, Neural Textures (NT) is the most challenging on both low and high compression due to highly realistic image synthesis. For a better understanding of the results, we visualize the regions of an image on which the network relies for classification using Grad-CAM \cite{Selvaraju_2019}. As shown in Fig. \ref{fig:fig3}, we observe that the proposed method focuses on a more localized region in both real and fake images as compared to the baseline method. This might be the result of better discriminative representations learned from both spatial and frequency domains.

\begin{table}[!tp]\centering
\scriptsize
\caption{\label{tab:tab2} Quantitative results (Acc \% and AUC) on the FaceForensics++ (FF++) dataset with high-quality HQ (or low compression c23) and low-quality LQ (or high compression c40) settings.}
\begin{tabular}{l|cc|ccc}\toprule
\multirow{2}{*}{Methods} &\multicolumn{2}{c|}{HQ} &\multicolumn{2}{c}{LQ} \\\cmidrule{2-5}
&Acc &AUC &Acc &AUC \\\midrule
Steg. Features \cite{6197267} &70.97 &- &55.98 &- \\
Cozzolino \emph{et al.} \cite{10.1145/3082031.3083247} &78.45 &- &58.69 &- \\
Rahmouni \emph{et al.} \cite{8267647} &79.08 &- &61.18 &- \\
Bayar and Stamm \cite{10.1145/2909827.2930786} &82.97 &- &66.84 &- \\
Face X-ray \cite{Li_2020_CVPR} &- &0.8740 &- &0.6160 \\
DSP-FWA \cite{li2019exposing} &- &0.5749 &- &0.6234 \\
MesoNET \cite{8630761} &83.10 &- &70.47 &- \\
Xception-ELA \cite{Xception-ELA} &93.86 &0.9480 &79.63 &0.8290 \\
FF++ \cite{Rossler_2019_ICCV} &95.04 &0.9509 &80.32 &0.8176 \\
Two-branch \cite{masi2020twobranch}&96.43 &0.9912 &86.34 &\underline{0.9110} \\
EfficientNet-B4 \cite{DBLP:conf/icml/TanL19} &96.63 &\underline{0.9918} &86.67 &0.8820 \\
$F^{3}$-Net \cite{Rossler_2019_ICCV} &\textbf{97.52} &0.9810 &\textbf{90.43} &\textbf{0.9330} \\
SPSL \cite{liu2021spatialphase} &91.50 &0.9532 &81.57 &0.8282 \\
\noalign{\vskip 0.2mm}
\hline
\noalign{\vskip 1mm}
Proposed MD-CSDNetwork &\underline{97.28} &\textbf{0.9929} &\underline{87.57} &0.8901 \\
\bottomrule
\end{tabular}
\end{table}

\subsubsection{Evaluation on Cross Databases}
It is necessary to evaluate the generalizability of the algorithm to reflect real-world scenarios. We used two large-scale datasets - Celeb-DF \cite{9156368} and DFD \cite{deepfakedetection}  for evaluating the transferability of our framework trained on FF++ \cite{Rossler_2019_ICCV}. This setting is more challenging since the testing set shares much less similarity with the training set. As shown in Table \ref{tab:tab3}, the proposed method shows better results when compared to several other previously published methods. Though, we still lag behind some state-of-the-art results \cite{Li_2020_CVPR,liu2021spatialphase, masi2020twobranch}, we obtain higher in-database results when compared with methods like \cite{Li_2020_CVPR,liu2021spatialphase} (refer Table \ref{tab:tab2}). The cross-database results on the DFD dataset are reported in Table \ref{tab:tab4}. The table shows that it performs significantly better than the baseline and comparable results with recent state-of-the-art methods.

\begin{table}[!tp]\centering
\scriptsize
\caption{\label{tab:tab3} Cross-dataset evaluation (AUC) on Celeb-DF \cite{9156368}. The model is trained on FF++ and tested on the Celeb-DF dataset. The results in the first column report the AUC values when tested only on the deepfake class in FF++. Our method outperforms most of the listed methods in cross-generalization. Results for other methods are directly cited from \cite{liu2021spatialphase}.}
\begin{tabular}{lccc}\toprule
Method &FF++ &Celeb-DF \\\midrule
Two-stream \cite{8014963} &0.7010 &0.5380 \\
Meso4 \cite{8630761} &0.8470 &0.5480 \\
Mesolnception4 \cite{8630761} &0.8300 &0.5360 \\
HeadPose \cite{8683164} &0.4730 &0.5460 \\
FWA \cite{li2019exposing} &0.8010 &0.5690 \\
VA-MLP \cite{8638330} &0.6640 &0.5500 \\
Xception-raw \cite{Rossler_2019_ICCV} &\textbf{0.9970} &0.4820 \\
Xception-c23 \cite{Rossler_2019_ICCV} &\textbf{0.9970} &0.6530 \\
Xception-c40 \cite{Rossler_2019_ICCV} &0.9550 &0.6550 \\
Multi-task \cite{9185974} &0.7630 &0.5430 \\
Capsule \cite{8682602} &0.9660 &0.5750 \\
DSP-FWA \cite{li2019exposing} &0.9300 &0.6460 \\
Face-XRay \cite{Li_2020_CVPR} &\underline{0.9912} &0.7420 \\
$F^{3}$-Net \cite{qian2020thinking} &0.9797 &0.6517 \\
Two-Branch \cite{masi2020twobranch} &0.9318 &\underline{0.7341} \\
EfficientNet-B4 \cite{DBLP:conf/icml/TanL19} &\textbf{0.9970} &0.6429 \\
SPSL \cite{liu2021spatialphase} &0.9691 &\textbf{0.7688} \\
\noalign{\vskip 0.2mm}
\hline
\noalign{\vskip 1mm}
Proposed MD-CSDNetwork &\textbf{0.9970} &0.6877 \\
\bottomrule
\end{tabular}
\end{table}

\begin{table}[!tp]\centering
\scriptsize
\caption{\label{tab:tab4} Cross-dataset evaluation (AUC) on DFD \cite{deepfakedetection}. The model is trained on FF++ and tested on the DFD dataset. The results for other methods are directly cited from \cite{luo2021generalizing}.}
\begin{tabular}{c|lll}\toprule
\multirow{5}{*}[-1em]{Trained on FF++} &Method &DFD \\\midrule
&Xception \cite{Chollet_2017_CVPR} &0.8310 \\
&Face X-Ray \cite{Li_2020_CVPR} &0.8560 \\
&High Frequency \cite{luo2021generalizing} &0.9190 \\
&Proposed MD-CSDNetwork &0.8951 \\
\bottomrule
\end{tabular}
\end{table}

\subsection{Ablation Study}

To understand the effectiveness of the proposed network, we quantitatively determine the importance of each component of the model. We evaluate the results for (a) baseline XceptionNet using only features from the spatial domain, (b) XceptionNet using only features from the frequency domain, (c) Combination of both without any cross stitches, (d) Combination of both with a single cross stitch, and (e) Combination of both with all the cross stitches. Table \ref{tab:tab5} shows the AUC results for each of these scenarios. We observe that using features only from the frequency domain does not perform well as compared to using features only from the spatial domain. Therefore, we cannot lose all the spatial information and rely completely on frequency domain features for classification. Instead, a combination of both the domains improves the performance significantly. A simple hard combination of features from both the domains by concatenating the last fully connected layers improves the AUC score. However in this case, there is no cross-talk between both the branches, and no layer parameters are shared. This gives limited room for information flow between both the domains. 

To see the effect of sharing any parameters at the intermediate layers, we first inserted only a single cross-stitch after the first max-pooling layer to connect the two branches. The applied stitch enables gradient flow at low-level feature maps and allows both direct and indirect supervision from the branches. An improved result as compared to the case of a simple concatenation of fully connected layers verifies our initial hypothesis of features from both the domains functioning as related supervisory signals helping one another for better generalization and performance. Next, we apply the remaining cross stitches after all the pooling layers in the network and observe further improvement in classification performance. This allows sharing at multiple levels in the network and ensures learning a better optimal combination of shared representations from both feature domains.

In another set of ablation experiments, we study the effect of using a different transformation in place of DCT for extracting the frequency domain features. We use the Discrete Wavelet Transform (DWT) \cite{wavelet} and the Fast Fourier Transform (FFT) \cite{bracewell1986fourier} as other types of image transformations. In FFT, the mathematical transformations applied to the input signal (or raw image) completely lose spatial information like DCT, whereas DWT offers a time-frequency representation of the image. For FFT, we use the amplitude spectrum as input frequency features in the network. We calculate the magnitudes of the series of complex numbers obtained as a result of the mathematical operations in FFT. As shown in Table \ref{tab:tab6}, the best results are observed with DCT. Using FFT, which outputs pure frequency domain features, yields lower but comparable results with DCT. Interestingly, DWT (Haar Wavelet Transform) gives the lowest results even though it contains both spatial and frequency information.

\begin{table}[!tp]\centering
\scriptsize
\caption{\label{tab:tab5} Ablation study and component analysis of the proposed method on high quality (HQ) setting in FF++. We add each component step-by-step and compare the obtained results in terms of the AUC values.}
\begin{tabular}{l|ccccc|cc}\toprule
ID &\vtop{\hbox{\strut RGB}\hbox{\strut Branch}}
&\vtop{\hbox{\strut Freq}\hbox{\strut Branch}} &\vtop{\hbox{\strut No}\hbox{\strut Stitches}}
&\vtop{\hbox{\strut One}\hbox{\strut Stitch}}
&\vtop{\hbox{\strut All}\hbox{\strut Stitches}} &AUC \\\midrule
1 &\checkmark & & & & &0.9509\\
2 & &\checkmark & & & &0.8114\\
3 &\checkmark &\checkmark &\checkmark & & &0.9789\\
4 &\checkmark &\checkmark & &\checkmark & &0.9900\\
5 &\checkmark &\checkmark & & &\checkmark &0.9929 \\
\bottomrule
\end{tabular}
\end{table}

\begin{table}[!tp]\centering
\scriptsize
\caption{\label{tab:tab6} Ablation study with different frequency transformations for extracting the frequency related features for our network. The table lists AUC values.}
\begin{tabular}{lcc}\toprule
Transform &AUC \\\midrule
DWT \cite{wavelet} &0.9665 \\
FFT \cite{bracewell1986fourier} &0.9800 \\
DCT \cite{1672377} &0.9929 \\
\bottomrule
\end{tabular}
\end{table}

\section{CONCLUSION and FUTURE WORK}
The paper presents MD-CSDNetwork, a novel approach of combining information obtained from spatial and frequency domains for improved face forgery detection. To learn this shared representation, we draw an analogy from the multi-task learning paradigm where multiple related supervisory tasks assist each other to learn better representations. We show that the features from two different domains (spatial and spectral) can also be treated as related supervisory training signals and an optimal shared representation can be learned via a cross-stitched network architecture. We obtain benchmark results for both intra-database and cross-database evaluation on challenging datasets. 

Generation and detection of deepfakes is like a cat and mouse game, and it requires continuously evolving detection methods as more efficient ways to create realistic media emerge. Current generation methods do not impose any constraints in the temporal dimension and may provide useful cues for improving the classification performance. In the future, we plan to explore inter-frame temporal artifacts to further improve the detection of deepfake videos.

\section*{Acknowledgements}
M. Vatsa is partially supported by the Department of Science and Technology, Government of India through Swarnajayanti Fellowship.

{\small
\bibliographystyle{ieee}
\bibliography{egbib}
}

\end{document}